\documentclass[10pt, conference]{IEEEtran}
\IEEEoverridecommandlockouts

\usepackage{cite}
\usepackage{amsmath,amssymb,amsfonts}
\usepackage{algorithmic}
\usepackage{graphicx}
\usepackage{textcomp}
\usepackage{xcolor}
\usepackage{url}
\usepackage{hyperref}
\usepackage{booktabs}
\usepackage{makecell}
\usepackage{svg}
\usepackage[export]{adjustbox}
\usepackage{enumitem}

\def\BibTeX{{\rm B\kern-.05em{\sc i\kern-.025em b}\kern-.08em
    T\kern-.1667em\lower.7ex\hbox{E}\kern-.125emX}}
\begin{document}

\title{Code Summarization Beyond Function Level}

\author{\IEEEauthorblockN{Vladimir Makharev}
\IEEEauthorblockA{	
\textit{Innopolis University, AIRI}\\
Innopolis, Russia \\
v.makharev@innopolis.university}
\and
\IEEEauthorblockN{Vladimir Ivanov}
\IEEEauthorblockA{\textit{Research Center of the Artificial Intelligence Institute}\\
\textit{Innopolis University}\\
Innopolis, Russia \\
v.ivanov@innopolis.ru}
}

\maketitle

\begin{abstract}
Code summarization is a critical task in natural language processing and software engineering, which aims to generate concise descriptions of source code. Recent advancements have improved the quality of these summaries, enhancing code readability and maintainability. However, the content of a repository or a class has not been considered in function code summarization. This study investigated the effectiveness of code summarization models beyond the function level, exploring the impact of class and repository contexts on the summary quality. The study involved revising benchmarks for evaluating models at class and repository levels, assessing baseline models, and evaluating LLMs with in-context learning to determine the enhancement of summary quality with additional context.
The findings revealed that the fine-tuned state-of-the-art \textit{CodeT5+ base} model excelled in code summarization, while incorporating few-shot learning and retrieved code chunks from RAG significantly enhanced the performance of LLMs in this task.
Notably, the \textit{Deepseek Coder 1.3B} and \textit{Starcoder2 15B} models demonstrated substantial improvements in metrics such as BLEURT, METEOR, and BLEU$_4$ at both class and repository levels. Repository-level summarization exhibited promising potential but necessitates significant computational resources and gains from the inclusion of structured context. Lastly, we employed the recent SIDE code summarization metric in our evaluation. This study contributes to refining strategies for prompt engineering, few-shot learning, and RAG, addressing gaps in benchmarks for code summarization at various levels. Finally, we publish all study details, code, datasets, and results of evaluation in the GitHub repository available at \url{https://github.com/kilimanj4r0/code-summarization-beyond-function-level}.
\end{abstract}


\vspace*{-2mm}
\section{Introduction}
\vspace*{-2mm}
Code summarization has emerged as a vital area in natural language processing (NLP) and software engineering (SE), aiming to generate concise and meaningful descriptions of source code. The ability to produce human-like summaries of code snippets have been significantly improved by advanced techniques leveraging neural networks, such as sequence-to-sequence models and Transformer-based architectures \cite{CodeBERT}, \cite{UniXcoder}. These advancements have led to the widespread adoption of automatic code summarization systems, enhancing code readability, maintainability, and overall understanding.

While considerable progress has been made in function-level code summarization, this focus neglects higher levels of abstraction such as classes and repositories. Indeed, summarizing code at the class and repository levels is crucial for comprehending complex and large-scale codebases, as such summarizing encompasses broader context and interactions within the software. Existing models often have limited capabilities in capturing the additional context provided by these higher-level structures, leading to a notable gap in effective techniques for summarizing complex codebases. For instance, when functions involve intricate structures and invoke others, repository-level context becomes essential.

\begin{figure}[!t]
    \begin{center}
        \includegraphics[width=0.75\linewidth]{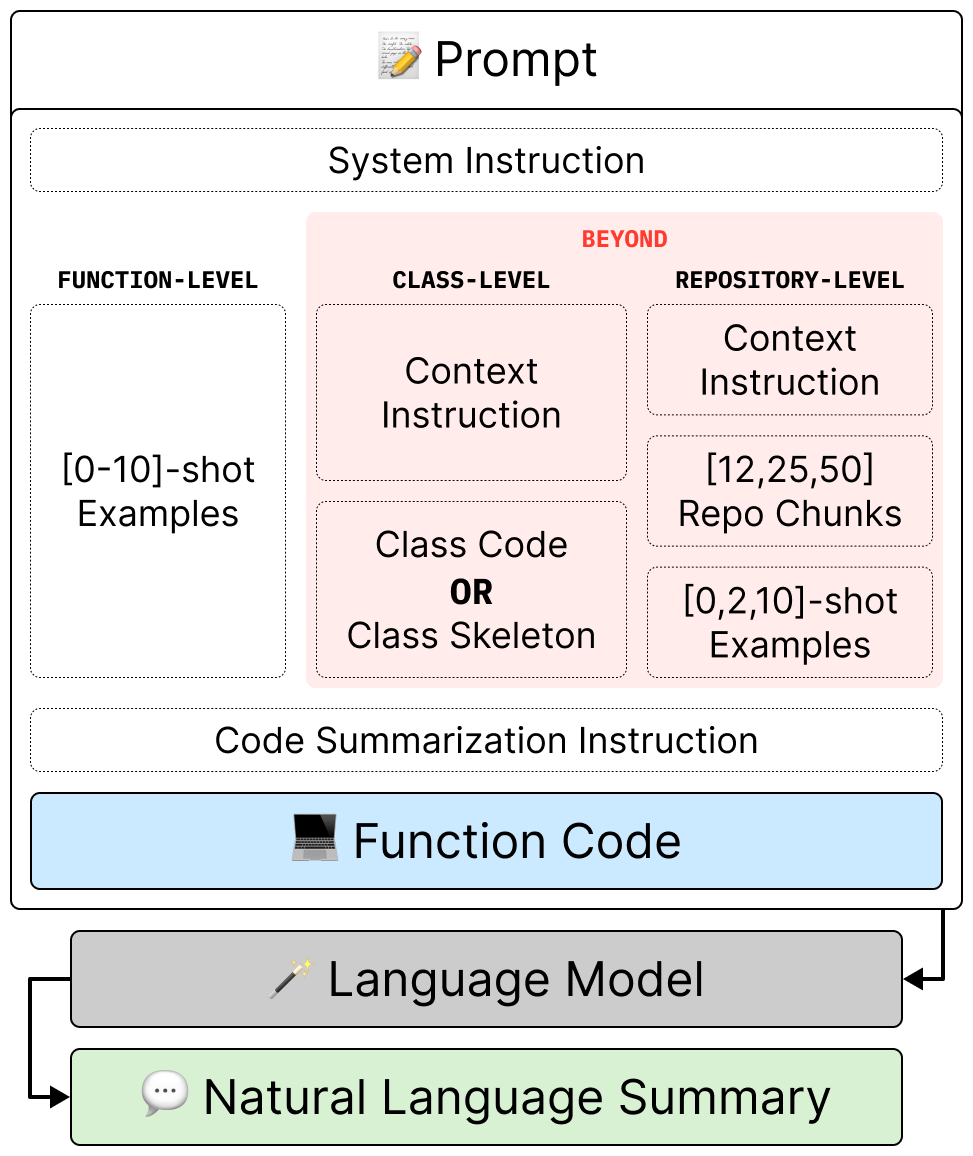}
    \end{center}
    \vspace*{-5mm}
    \caption{Code summarization pipeline at function, class, and repository levels, as used in this study.}
    \label{fig:code-summarization-schema}
    \vspace*{-7mm}
\end{figure}

Addressing this gap is critical for advancing the field and enhancing the utility of code summarization in real-world applications. Incorporating class and repository contexts can potentially provide more comprehensive and meaningful summaries, capturing essential functional and architectural details for developers. Moreover, code summarization models remain insufficiently evaluated beyond individual functions.

This study aims to explore the effectiveness of code summarization models beyond the function level by investigating the impact of additional context from classes and repositories. Fig. \ref{fig:code-summarization-schema} presents the overall code summarization pipeline schema used in this work. We hypothesize that this broader context would enrich source code, especially when combined with the utilization of Large Language Models (LLMs) and in-context learning techniques, such as Retrieval-augmented generation (RAG) \cite{RAGSurvey}. Such enrichment will obviously improve the quality of automatically generated code summaries. The additional context is expected to enable models to produce more comprehensive and concise summaries, enhancing code understanding and maintainability.

To test this hypothesis, we evaluate various code summarization models, including both pre-trained language models and LLMs with in-context learning capabilities, using revised benchmarks that assess performance at function, class, and repository levels. During the evaluation, we used six metrics to analyze the results and determine the broader context effectiveness in code summarization. Our findings revealed improved code summarization quality with the inclusion of additional context and few-shot learning, highlighting the potential of LLMs in this domain.

\vspace*{-1mm}
\section{Methodology}
\label{methodology}
\vspace*{-1mm}

This section presents the methodology used to evaluate the effectiveness of code summarization models beyond the function level. Our study addressed a significant gap in the literature by integrating class and repository-level information into code summarization. We aimed to enhance the accuracy and comprehensiveness of automated code summarization techniques. To investigate the research question, \textbf{"How effective are code summarization models beyond the function level?"}
, we conducted experiments comparing state-of-the-art models at the function, class, and repository levels. The evaluation used widely accepted text generation metrics on benchmark datasets.

For simplicity, we treated methods and functions as equivalent blocks. \textit{Python} was the sole programming language used in this study. A "summary" refers to a concise \textit{English} description of a function code snippet, typically consisting of 1-3 sentences. We chose Python and English due to their widespread use in software development and prevalence in benchmarks, enhancing the relevance and understanding of our research outcomes. Function-level code summarization focused on summarizing individual functions, while class-level and repository-level code summarization provided additional context by summarizing functions within classes and repositories, respectively.

\vspace*{-1mm}
\subsection{Evaluation Datasets}

To evaluate code summarization at function, class, and repository levels, we used: the Modified ClassEval dataset and the Modified CodeSearchNet dataset. These datasets are based on the ClassEval benchmark \cite{du2023classeval} and the CodeSearchNet dataset from the CodeXGLUE benchmark \cite{CodeXGLUE}, respectively.

\begin{figure}[h]
    \begin{center}
        \includegraphics[width=0.75\linewidth]{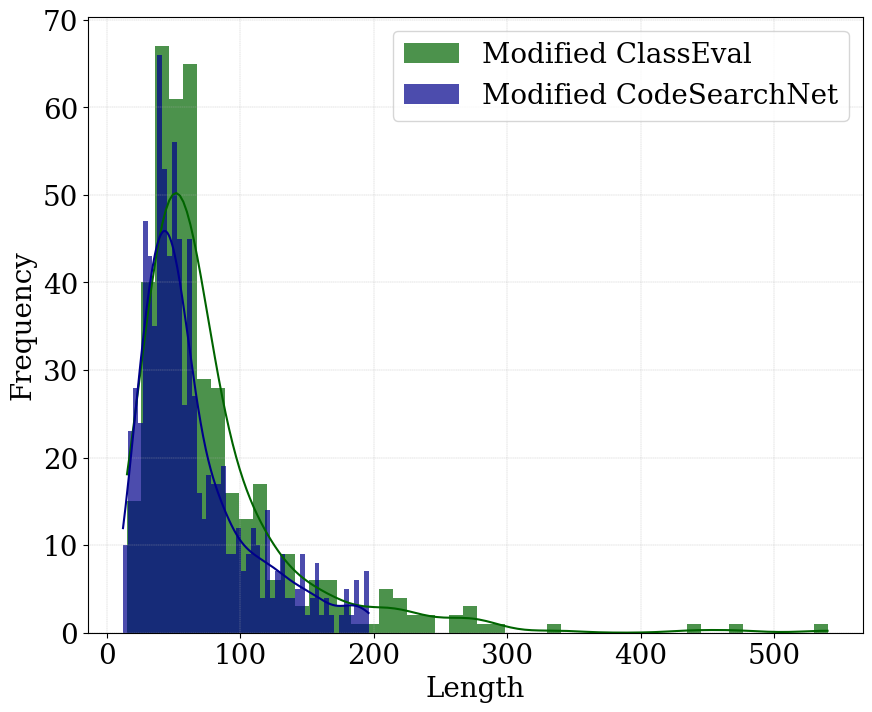}
    \end{center}
    \vspace*{-5mm}
    \caption{Function summary length distributions of modified ClassEval and CodeSearchNet datasets.}
    \label{fig:datasets_distribution}
    \vspace*{-5mm}
\end{figure}

Both datasets were obtained using the HuggingFace \texttt{datasets} Python package due to its convenient interface. Modifications applied to the datasets are summarized in Table \ref{tab:eval_datasets}. As a result, we obtained modified datasets with comparable function summary length distributions depicted in Fig. \ref{fig:datasets_distribution}.


\begin{table}[!h]
\vspace*{-3mm}
\caption{Evaluation Datasets Modification}
\label{tab:eval_datasets}
\vspace*{-2mm}
\centering
\footnotesize
\begin{tabular}{lcc}
\toprule
\textbf{Dataset} & \textbf{ClassEval} & \textbf{CodeSearchNet} \\
\midrule
Functions & 410 & 14,918 \\
\midrule
\multicolumn{3}{l}{\makecell[l]{\textit{After modifications: extracting, filtering, and cleaning}}} \\
\toprule
Functions for Evaluation & 400 & 806 \\
Functions for Few-Shot Learning & 10 & 40 (10 per 4 repos) \\
\bottomrule
\end{tabular}
\vspace*{-3mm}
\end{table}

\subsubsection{Modified ClassEval}

The Modified ClassEval dataset is derived from the ClassEval benchmark \cite{du2023classeval}, a class-level Python code generation benchmark. This benchmark comprises 100 manually crafted class-level coding tasks, covering 100 classes and 410 methods. We extracted tuples of (context, function code, function summary) for each Python class. The context could be one of the following: (1) empty; (2) class code without the function code to be summarized; (3) class skeleton without the function code to be summarized. Regular expressions were employed to extract summaries from function descriptions and to extract function code to be summarized from class code and skeletons. The distribution in Fig. \ref{fig:datasets_distribution} shows that most function summaries were no longer than 125 characters.

Next, we randomly selected ten tuples for experiments involving few-shot learning. As a result, we obtained three subsets of the Modified ClassEval dataset, each formatted in JSONL and comprising 400 and 10 tuples of (context, function code, function summary) for evaluation and few-shot learning, respectively.

\subsubsection{Modified CodeSearchNet}

The Modified CodeSearchNet dataset is based on the CodeSearchNet dataset from the CodeXGLUE benchmark \cite{CodeXGLUE}, which comprises one million filtered pairs of (comment, code) extracted from open-source repositories. We selected the test split comprising 14,918 tuples of (repository name, function code, function documentation) from the Python subset.

Initially, we selected the top repositories from GitHub using the GitHub API, resulting in four repositories after refinement of the selection.
Table \ref{tab:filtered_repos} presents selected repositories. We employed regular expressions to extract summaries from function documentation and remove documentation and comments from function codes. We then eliminated function summaries shorter than ten or longer than 200 characters, leading to the distribution depicted in Fig. \ref{fig:datasets_distribution}.

For each repository, we selected 846 tuples: 806 for evaluation and 40 for few-shot learning (10 per repository). The modified dataset consisted of tuples comprising (repository name, function code, function summary). This modification facilitated incorporating repository context into repository-level code summarization.


\begin{table*}[h]
\vspace*{-3mm}
\centering
\caption{Filtered Repositories for Modified CodeSearchNet Dataset (as of April 26, 2024)}
\label{tab:filtered_repos}
\vspace*{-2mm}
\begin{tabular}{lccccc}
\toprule
\textbf{Repository} & \textbf{Stars} & \textbf{Forks} & \textbf{Functions Extracted} & \textbf{Avg Chunk Size (Chars)} & \textbf{Chunks} \\
\midrule
\href{https://github.com/apache/airflow}{apache/airflow} & 34,500 & 13,545 & 455 & 818 & 42,925 \\
\href{https://github.com/streamlink/streamlink}{streamlink/streamlink} & 9,564 & 1,074 & 64 & 804 & 3,328 \\
\href{https://github.com/open-mmlab/mmcv}{open-mmlab/mmcv} & 5,601 & 1,587 & 50 & 771 & 1,940 \\
\href{https://github.com/Azure/azure-sdk-for-python}{Azure/azure-sdk-for-python} & 4,275 & 2,672 & 237 & 839 & 1,209,693 \\
\bottomrule
\end{tabular}
\vspace*{-5mm}
\end{table*}

\vspace*{-1mm}
\subsection{Models}
\label{2.B}

We evaluated five baseline language models and five LLMs on the modified datasets (see Table \ref{tab:models_used}). These models were chosen for their distinct architectures, relevance to code summarization tasks, and availability within the Hugging Face model hub.

\begin{table*}[!t]
\vspace*{-3mm}
\caption{Models Used in the Study}
\label{tab:models_used}
\vspace*{-2mm}
\centering
\footnotesize
\begin{tabular}{lclcr}
\toprule
\textbf{Model} & \textbf{Reference} & \textbf{Shorten Name} & \textbf{Parameters} & \textbf{Memory Usage} \\
\midrule
\textit{\textbf{Baseline Language Models}} & & & & \\
\toprule
\href{https://hf.co/SEBIS/code_trans_t5_large_source_code_summarization_python_multitask_finetune}{SEBIS/code\_trans\_t5\_large\_source\_code\_sum...} & \cite{CodeTrans} & ct-t5-large-sum & 770M & 2.75 GB \\
\href{https://hf.co/SEBIS/code_trans_t5_large_code_documentation_generation_python_multitask_finetune}{SEBIS/code\_trans\_t5\_large\_code\_doc...} & \cite{CodeTrans} & ct-t5-large-doc & 770M & 2.75 GB \\
\href{https://hf.co/Salesforce/codet5-base-multi-sum}{Salesforce/codet5-base-multi-sum} & \cite{CodeT5} & codet5-base & 220M & 850 MB \\
\href{https://hf.co/Paul-B98/codet5p_220m_py_sum}{Paul-B98/codet5p\_220m\_py\_sum} & \cite{CodeT5p} & codet5p-base & 220M & 850 MB \\
\href{https://hf.co/lintang/pile-t5-large-codexglue}{lintang/pile-t5-large-codexglue} & \cite{2024PileT5} & pile-t5-large & 783M & 2.79 GB \\
\toprule
\textit{\textbf{Large Language Models}} & & & & \\
\toprule
\href{https://hf.co/deepseek-ai/deepseek-coder-1.3b-instruct}{deepseek-ai/deepseek-coder-1.3b-instruct} & \cite{DeepSeekCoder} & deepseek-coder-1.3b & 1.3B & 2.57 GB \\
\href{https://hf.co/deepseek-ai/deepseek-coder-6.7b-instruct}{deepseek-ai/deepseek-coder-6.7b-instruct} & \cite{DeepSeekCoder} & deepseek-coder-6.7b & 6.7B & 13.75 GB \\
\href{https://hf.co/deepseek-ai/deepseek-coder-33b-instruct}{deepseek-ai/deepseek-coder-33b-instruct} & \cite{DeepSeekCoder} & deepseek-coder-33b & 33B & 62.16 GB \\
\href{https://hf.co/bigcode/starcoder2-15b-instruct-v0.1}{bigcode/starcoder2-15b-instruct-v0.1} & \cite{StarCoder2} & starcoder2-15b & 15B & 29.47 GB \\
\href{https://hf.co/gradientai/Llama-3-8B-Instruct-Gradient-1048k}{gradientai/Llama-3-8B-Instruct-Gradient-1048k} & \cite{Llama3} & llama3-8b & 8B & 29.98 GB \\
\bottomrule
\end{tabular}
\vspace*{-6mm}
\end{table*}

\subsubsection{Baseline Language Models}

We selected five baseline language models specifically fine-tuned for code summarization tasks. The first two models were based on the CodeTrans \cite{CodeTrans} architecture: \textit{ct-t5-large-sum} and \textit{ct-t5-large-doc}. The \textit{ct-t5-large-sum} model was fine-tuned to generate concise summaries for Python code snippets, while \textit{ct-t5-large-doc} was optimized for generating detailed documentation from Python code. Both models leveraged the T5-large architecture to capture Python's syntactic and semantic nuances.

The next two models, \textit{codet5-base} and \textit{codet5p-base}, were based on the CodeT5 \cite{CodeT5} and CodeT5+ \cite{CodeT5p} architectures. \textit{Codet5-base} was designed for multi-task code summarization across various programming languages, demonstrating proficiency in generating summaries for diverse code snippets. \textit{Codet5p-base} was specifically trained for Python code summarization tasks, potentially capturing intricate patterns unique to Python code due to its specialized training.

The last model was \textit{pile-t5-large} recently introduced by Sutawika et al. \cite{2024PileT5}. The authors refined the tokenizers used in standard T5 models by adapting them with the LLaMA tokenizer \cite{LLaMATokenizer} to improve handling of code-specific syntax, variable names, and structural tokens, ensuring better compatibility with code tokens. The \textit{pile-t5-large} model leverages the T5 architecture fine-tuned on the Pile dataset for code-to-text generation tasks, including code summarization. Trained on diverse code snippets, this model generalizes well across different codebases.

For inference, the baseline models were run using the HuggingFace \texttt{transformers} library. We improved the quality and diversity of generated summaries through the \texttt{SummarizationPipeline} with a beam search multinomial sampling approach, integrating randomness with structured guidance. A beam search with five beams and multinomial sampling outperformed the greedy approach in initial experiments by generating more diverse and contextually accurate code summaries.

Inference was conducted for each baseline model over both evaluation datasets at the function level. This setup allowed us to assess the performance of models specifically fine-tuned for code summarization tasks.

\subsubsection{Large Language Models}

We selected five LLMs to evaluate their performance in code summarization tasks. The DeepSeek Coder series \cite{DeepSeekCoder}, comprising \textit{deepseek-coder-1.3b}, \textit{deepseek-coder-6.7b}, and \textit{deepseek-coder-33b}, are cutting-edge open-source models tailored for coding across diverse programming languages. These models leverage an extensive pretraining dataset of 2 trillion tokens, predominantly code data, enabling superior performance in code understanding and generation tasks. The different model sizes allow us to evaluate performance across diverse computational resources.

The \textit{starcoder2-15b} model was designed specifically for code-related tasks, utilizing the recent StarCoder2 \cite{StarCoder2} architecture. \textit{starcoder2-15b} was the first entirely self-aligned code model trained without human annotations or proprietary data. With 15 billion parameters, this model can respond to coding-related instructions in multiple programming languages, making it suitable for code-related tasks.

Lastly, the \textit{llama3-8b} model is based on the LLaMA 3 \cite{Llama3} architecture and supports extended context lengths exceeding 1 million tokens. \textit{llama3-8b} is one of the most recent open-source LLMs, that processes long contexts more effectively compared to other models such as Mistral \cite{Mistral} and Gemma \cite{Gemma}. This feature highlights its potential for enhancing code summarization at the repository level, where long-range context is beneficial.

Notably, the \textit{llama3-8b} model exhibits a substantially larger memory footprint (see Table \ref{tab:models_used}), nearly doubling that of its base version with an 8K context length. This doubling is due to its ability to support extended context lengths exceeding 1 million tokens, which is beneficial for repository-level code summarization but requires more computational resources. All models were evaluated using half-precision floating point (\texttt{float16}) data types to optimize memory usage without significant loss of precision.

For inference, each LLM was run using the HuggingFace \texttt{transformers} library with individual generation configurations. As advised by the authors of the selected models, a greedy decoding approach was uniformly selected for all models. The maximum output length was set at 128 tokens to encompass the majority of summary lengths in the evaluation datasets and to decrease generation time. This configuration promoted consistency and determinism in model output, optimizing efficiency by reducing computational overhead.

Inference of each LLM was conducted at the function, class, and repository levels with varying generation configurations. We used the following context windows to enrich instructions with varying lengths of context: 1M for Llama 3 and 16K for DeepSeek Coder series and StarCoder2. Each LLM was evaluated for function-level code summarization with 1-10 few-shot examples on both evaluation datasets. Subsequently, using the Modified ClassEval dataset, each LLM was assessed for class-level code summarization, with class code or skeleton serving as context. Finally, using the Modified CodeSearchNet dataset, each LLM was integrated into the Naive RAG to evaluate its performance at the repository level.

\vspace*{-1mm}
\subsection{Evaluation}

We assessed the quality of the generated code summaries, with the following six widely accepted metrics: BLEU$_4$ \cite{BLEU}, ROUGE$_L$ \cite{ROUGE}, METEOR \cite{METEOR}, BERTScore$_{F1}$ \cite{BERTScore}, BLEURT \cite{BLEURT}, and SIDE \cite{SIDE}. These metrics were computed using the HuggingFace \texttt{evaluate} library.

\begin{itemize}[leftmargin=3mm]
    \item \textit{BLEU$_4$}: Calculated the similarity between the generated and reference summaries based on 1- to 4-gram precision, adjusted by a brevity penalty for shorter sentences.
    \item \textit{ROUGE$_L$}: Assessed the structural similarity by identifying the longest common subsequence between the generated and reference summaries, considering both precision and recall.
    \item \textit{METEOR}: Emphasized recall over precision using the harmonic mean of unigram precision and recall, incorporating synonyms and stemming variations.
    \item \textit{BERTScore$_{F1}$}: Evaluated semantic similarity using embeddings from a pre-trained BERT model, capturing contextual variability in language and combining precision and recall.
    \item \textit{BLEURT}: A learned evaluation metric that used transfer learning and human annotations to capture semantic and pragmatic aspects of the summaries.
    \item \textit{SIDE}: A metric tailored for code summarization that modeled the characteristics of suitable and unsuitable code summaries using contrastive learning.
\end{itemize}

For BERTScore$_{F1}$, BLEURT, and SIDE the top-performing neural networks recommended by their respective authors: \textit{microsoft/deberta-xlarge-mnli\footnote{\url{https://github.com/Tiiiger/bert_score}}}, \textit{BLEURT-20\footnote{\url{https://github.com/google-research/bleurt}}}, \textit{side-hard-negatives-fine-tuned\footnote{\url{https://github.com/antonio-mastropaolo/code-summarization-metric}}}, respectively.

In our experiments, the metric results of each model were averaged for each level in each dataset. For the repository level, we also computed the average metrics for each repository project separately. This comprehensive evaluation allows us to assess the performance of different models across various code summarization tasks.

The comparison of code summarization experiments' results was structured as follows:
\begin{itemize}[leftmargin=3mm]
    \item Between baselines and LLMs at the function level.
    \item Between LLMs at the function level with and without 1-10 few-shot examples, at the class level with class code or skeleton, and at the repository level with and without the optimal number of few-shot examples, using varying lengths of context.
    \item Between LLMs per repository project at the function level and repository level.
\end{itemize}

This comparison procedure allows for a detailed examination of code summarization performance across various levels, providing insights into the efficacy of different models. Furthermore, Table \ref{tab:models_used} introduces shortened names for each chosen model for improved readability.

All experiments were conducted using two NVIDIA A100 GPUs with 80 GB of memory each. The high computational capacity of these GPUs enabled efficient processing of large models and datasets, ensuring that resource constraints did not impede the evaluation.

\vspace*{-1mm}
\section{Experiment Design}
\vspace*{-1mm}

\begin{figure*}[!t]
    \begin{center}
        \includegraphics[width=0.7\textwidth]{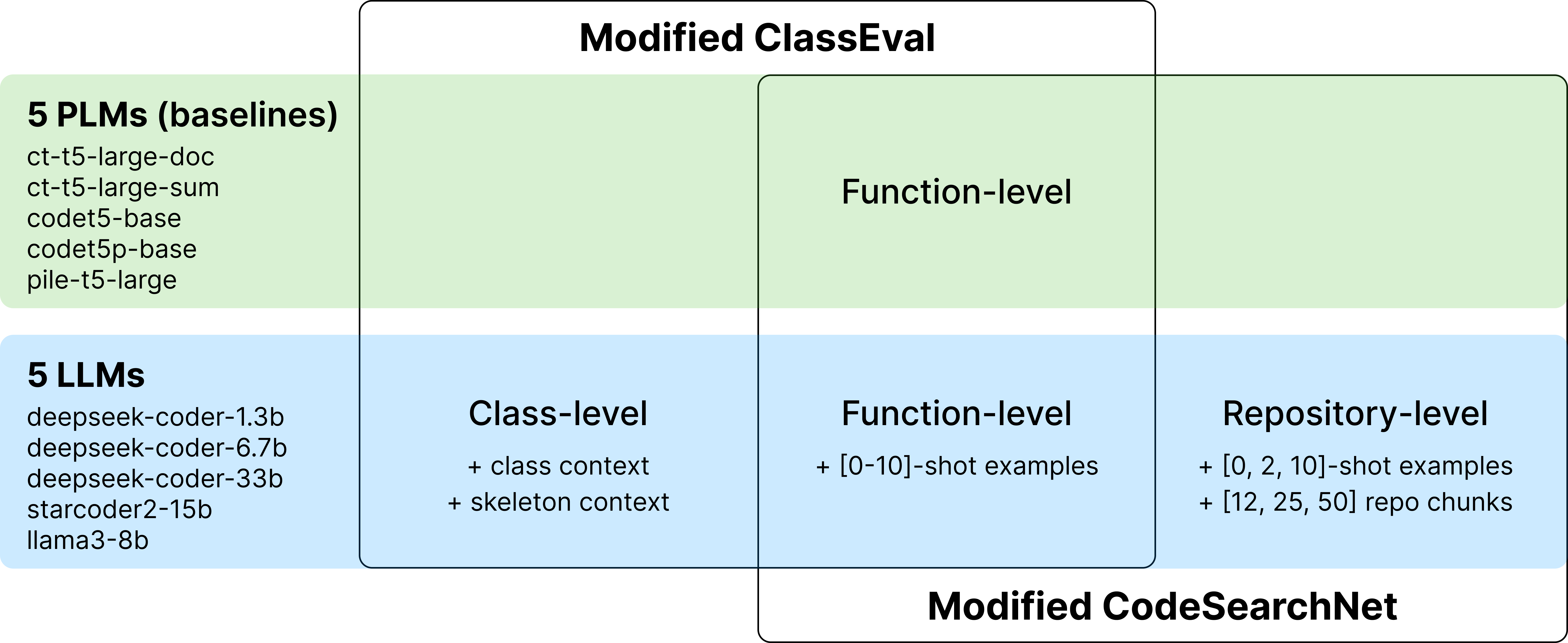}
    \end{center}
    \vspace*{-4mm}
    \caption{Experiments conducted in this study.}
    \label{fig:experiments}
\vspace*{-6mm}
\end{figure*}

We conducted experiments to evaluate code summarization models at the function, class, and repository levels using the methods previously described. Fig. \ref{fig:experiments} overviews these experiments, most of which involved LLMs. To facilitate reproducibility and contribute to the research community, we published the Modified ClassEval\footnote{\url{https://hf.co/datasets/sm1rk/modified-classeval-code-summarization}} and Modified CodeSearchNet\footnote{\url{https://hf.co/datasets/sm1rk/modified-codesearchnet-code-summarization}} datasets on the HuggingFace Hub.

During the experiments, we employed the inference procedure outlined in Section \ref{2.B}. Fig. \ref{fig:code-summarization-schema} illustrates the code summarization pipeline schema employing a structured prompt with varying levels of context. For the code summarization task, baseline models received only the function code as input, while LLMs were provided with a prompt that included the function code and additional context. The expected output from all models was a concise summary of the function code. To guide the LLMs toward producing relevant summaries, we crafted \textbf{a custom system prompt}:
\vspace*{-2mm}
\begin{verbatim}
You're a specialized AI assisting with 
Python code summaries, deeply
knowledgeable in computer science.
\end{verbatim}
This prompt was designed to focus the models on generating succinct and accurate code summaries. All prompts in our experiments are provided in detail in the paper’s GitHub repository. The most relevant components of these prompts are highlighted below.

\vspace*{-1mm}
\subsection{Function-level Experiments}

Function-level code summarization experiments involved both baseline models and LLMs, evaluated using function code as input. Baseline models were assessed solely on this code, while LLMs were tested in few-shot setting, including zero-shot. In the zero-shot setup, LLMs received only the function code and a system prompt without prior examples. In the few-shot setup, 1–10 reference examples were provided as prior message history before the code summarization instruction, guiding the LLM through illustrative cases. \textbf{The few-shot examples template} was as follows:
\vspace*{-3mm}
\begin{verbatim}
{
<few-shot function code>
Concisely summarize the Python code
provided in 1-3 sentences.
<few-shot summary>
} * [1-10]-shot examples
\end{verbatim}
\vspace*{-2mm}
The main finding was that LLMs showed consistent improvements as the number of few-shot examples increased. Despite advancements in LLMs, baseline models remained superior in this context. 


\vspace*{-1mm}
\subsection{Class-level Experiments}

Class-level code summarization experiments focused on evaluating LLMs using either the full class code or a class skeleton\footnote{The class skeleton provides a structured blueprint for the target class, including class-level and function-level information as defined in \cite{du2023classeval}.} as context. We designed \textbf{a class context instruction template} to incorporate this additional information: 
\vspace*{-3mm}
\begin{verbatim}
Consider the following class code as
additional context for your response:
<class code or class skeleton>
\end{verbatim}
\vspace*{-2mm}
Our findings demonstrated that including the skeleton context consistently improved the performance of the LLMs compared to using the full class code as context.


\subsection{Repository-level experiments}
\label{3.C}

In our repository-level code summarization experiments, we used LLMs integrated within a Naive RAG pipeline to handle the broader context of entire code repositories. The LLMs were provided with several few-shot examples and code chunks retrieved from the repository as context. Note that some of the code chunks included docstrings though not the target code summary. However, these docstrings did not serve the same purpose as few-shot examples. Few-shot examples were explicitly chosen to demonstrate the structure or style of desired outputs over multiple turns, effectively simulating a conversation or iterative context for the model. By contrast, the docstrings within chunks were simply part of the contextual information provided in a single prompt and did not replicate the guiding role that few-shot examples fulfill.

We implemented a pipeline that included downloading repositories, filtering for Python files, chunking the code into manageable pieces, and constructing a FAISS \cite{FAISS} index to facilitate efficient similarity search. The code chunks were embedded using a high-performing embedding model \texttt{BAAI/bge-large-en-v1.5}\footnote{\url{https://hf.co/BAAI/bge-large-en-v1.5}}
chosen for its optimal size and high-quality English language support. We selected the top-\(K\) most similar chunks based on cosine similarity as context for the LLMs, with \(K\) set to 12, 25, and 50. These values were selected to fully utilize the LLM's 16K context window of approximately 50 chunks or 15K tokens, with \(K\) halved to facilitate comparison.

\textbf{A repository context instruction template} was designed to present retrieved code chunks with their file paths and content:
\vspace*{-5mm}
\begin{verbatim}
You have the following repository context,
which includes fragments of code with
their corresponding paths and lines from 
the repository:
{
File path: <path of file with code chunk>
File content:
```
<code chunk content>
```
} * [12, 25, 50] repo chunks
\end{verbatim}

This instruction simulated a developer's need to understand a specific Python function within a repository context. The code summarization instruction asked the LLM to summarize the function based on this context, exploring the model's ability to handle large-scale codebases and generate accurate summaries informed by relevant information.

The experiment results indicated that while including code chunks as context did not significantly affect the performance of LLMs when used alone, introducing few-shot examples alongside the context led to noticeable improvements.


\vspace*{-1mm}
\section{Results Analysis and Discussion}
\vspace*{-1mm}

The key results are summarized in Table \ref{tab:results-union}, highlighting the effectiveness of code summarization models beyond the function level.
We compared the highest-performing baseline model, as determined by metric values, with LLMs in several configurations across selected datasets and levels. Note that running baseline models at the class or repository level yielded meaningless results because these models were trained solely on function code, without natural language instructions as prompts.

The baseline model, \textit{codet5p-base}, demonstrated strong performance due to its state-of-the-art architecture and fine-tuning on the filtered CodeSearchNet dataset \cite{CodeSearchNet}. When LLMs were provided with few-shot examples or additional context, they showed significant improvements in generating concise and relevant code summaries.

Including few-shot examples is crucial for enhancing the performance of LLMs in code summarization, as these examples guide LLMs to produce more concise and aligned summaries, reducing their tendency to produce overly explanatory outputs without such guidance. Comparative analysis showed that LLMs with few-shot examples could match or even surpass baseline models in certain metrics, especially when the examples were of high quality. While baseline models retained advantages in certain areas due to fine-tuning, LLMs' adaptability with minimal examples underscored their potential for code summarization beyond the function level. However, the success of LLMs depended on choosing the right examples and providing relevant context.

\begin{table*}[!t]
\vspace*{-3mm}
\centering
\caption{Code Summarization Results on Modified ClassEval and Modified CodeSearchNet Datasets}
\label{tab:results-union}
\vspace*{-2mm}
\begin{tabular}{lllcccccc}
\toprule
\textbf{Dataset} & \textbf{Model} & \textbf{Configuration} & \textbf{SIDE}\small{$\downarrow$} & \textbf{BLEURT} & \textbf{BERTScore$_{F1}$} & \textbf{ROUGE$_{L}$} & \textbf{METEOR} & \textbf{BLEU$_{4}$} \\
\midrule
\multicolumn{2}{l}{\textit{\textbf{Modified ClassEval}}} & & & & & & & \\
\toprule
function-level & codet5p-base & baseline & \textbf{.240} & .534 & .726 & 41.43 & 35.38 & 8.55 \\
& deepseek-coder-1.3b & LLM + 0-shot & .719 & .549 & .640 & 26.04 & 36.57 & 3.83 \\
& deepseek-coder-1.3b & LLM + 10-shot & .412 & .611 & .747 & 42.80 & 42.82 & 11.69 \\
& llama3-8b & LLM + 10-shot & .365 & \textbf{.620} & .747 & 42.37 & 43.20 & 11.99 \\
& starcoder2-15b & LLM + 8-shot & .269 & .616 & \textbf{.757} & \textbf{43.62} & 42.40 & \textbf{12.62} \\
\hline
class-level & deepseek-coder-1.3b & LLM + class context & .693 & .576 & .684 & 36.85 & 46.82 & 9.02 \\
& deepseek-coder-1.3b & LLM + skeleton context & .676 & .586 & .691 & 36.74 & \textbf{47.87} & 10.50 \\
& deepseek-coder-6.7b & LLM + skeleton context & .882 & .544 & .625 & 25.85 & 39.72 & 5.67 \\
& deepseek-coder-33b & LLM + skeleton context & .934 & .536 & .605 & 24.34 & 38.31 & 5.10 \\
& llama3-8b & LLM + skeleton context & .834 & .560 & .629 & 25.81 & 41.27 & 6.46 \\
& starcoder2-15b & LLM + skeleton context & .937 & .551 & .562 & 21.33 & 34.02 & 4.03 \\
\toprule
\multicolumn{2}{l}{\textit{\textbf{Modified CodeSearchNet}}} & & & & & & & \\
\toprule
function-level & codet5p-base & baseline & .075 & .502 & .721 & 35.68 & 33.63 & 4.30 \\
& deepseek-coder-1.3b & LLM + 0-shot & .659 & .481 & .587 & 17.07 & 28.29 & 0.79 \\
& deepseek-coder-1.3b & LLM + 2-shot & .142 & .529 & .706 & 32.05 & 35.24 & 3.26 \\
& deepseek-coder-1.3b & LLM + 10-shot & .068 & .539 & .724 & 34.30 & 34.48 & 4.46 \\
& llama3-8b & LLM + 10-shot & \textbf{.041} & .545 & .728 & 34.60 & 34.99 & 5.33 \\
& starcoder2-15b & LLM + 10-shot & .081 & .560 & \textbf{.738} & \textbf{38.15} & 37.67 & 7.10 \\
\hline
repository-level & deepseek-coder-1.3b & LLM + 2-shot + 12 chunks & .277 & .566 & .699 & 35.42 & \textbf{44.08} & 9.70 \\
& deepseek-coder-1.3b & LLM + 0-shot + 12 chunks & .813 & .479 & .562 & 20.50 & 36.40 & 3.10 \\
& deepseek-coder-6.7b & LLM + 10-shot + 50 chunks & .241 & \textbf{.568} & .706 & 36.82 & 43.65 & \textbf{11.89} \\
& deepseek-coder-33b & LLM + 10-shot + 25 chunks & .568 & .511 & .626 & 22.60 & 35.90 & 3.40 \\
& deepseek-coder-33b & LLM + 10-shot + 50 chunks & .594 & .510 & .622 & 22.00 & 35.40 & 3.10 \\
& llama3-8b & LLM + 10-shot + 50 chunks & .511 & .532 & .624 & 24.80 & 38.30 & 6.70 \\
& starcoder2-15b & LLM + 10-shot + 50 chunks & .643 & .506 & .582 & 18.50 & 29.50 & 2.70 \\
\midrule
\multicolumn{9}{l}{Ground truth value of SIDE for Modified ClassEval is 0.476, for Modified CodeSearchNet is 0.175.} \\
\multicolumn{9}{l}{\textit{Detailed evaluation results for all experiments are available in the GitHub repository for this study.}}
\end{tabular}
\vspace*{-6mm}
\end{table*}

\vspace*{-1mm}
\subsection{Discussion of specific results}

This subsection examines the detailed outcomes across different datasets and levels, emphasizing how the incorporation of contextual information and few-shot learning influences the performance of LLMs compared to the highest-performing baseline model.

In the Modified ClassEval dataset, the baseline model \textit{codet5p-base} achieved a SIDE score of 0.240 at the function level, demonstrating strong alignment with the ground truth summaries. LLMs in zero-shot configurations exhibited higher SIDE scores, indicating less similarity to the ground truth. An example of such LLMs is \textit{deepseek-coder-1.3b}. However, when provided with few-shot examples, LLMs significantly improved their performance. The \textit{starcoder2-15b} model, with 8-shot examples, achieved the highest scores in BERTScore$_{F1}$ (0.757), ROUGE$_{L}$ (43.62), and BLEU$_{4}$ (12.62), outperforming the baseline and other LLMs in these evaluation metrics.

At the class level, the \textit{deepseek-coder-1.3b} model with skeleton context had a higher SIDE score of 0.676, suggesting greater divergence from the ground truth compared to function-level summaries. This unexpected result indicates that simply providing class context without guiding examples may not lead to more concise summaries. The importance of few-shot examples is further emphasized, as models tend to benefit more from structured guidance than from broader context alone. Overall, the Modified ClassEval results demonstrated the efficacy of few-shot learning in improving LLM performance for code summarization tasks. Notably, using only the skeleton structure outperformed entire classes, likely because full class context introduced too much noise, making it harder for the model to extract relevant information for concise summaries.

In the Modified CodeSearchNet dataset, the baseline model \textit{codet5p-base} maintained a competitive SIDE score of 0.075 at the function level. LLMs in zero-shot configurations performed poorly, with \textit{deepseek-coder-1.3b} showing a high SIDE score of 0.659. However, with the inclusion of 10-shot examples, \textit{llama3-8b} and \textit{starcoder2-15b} showed significant improvements. The \textit{llama3-8b} model achieved the lowest SIDE score of 0.041, closely matching the baseline, while \textit{starcoder2-15b} excelled in BERTScore$_{F1}$ (0.738), ROUGE$_{L}$ (38.15), and BLEU$_{4}$ (7.10), indicating superior alignment with the ground truth in these metrics.

At the repository level, LLMs equipped with few-shot examples and additional context outperformed function-level models in certain metrics. The \textit{deepseek-coder-6.7b} model was configured with 10-shot examples and 50 chunks of code context, which is the maximum that filled the LLM's context window. This \textit{deepseek-coder-6.7b} model achieved higher BLEURT (0.568) and BLEU$_{4}$ (11.89) scores compared to function-level results. Similarly, the smaller \textit{deepseek-coder-1.3b} model was configured with 10-shot examples and 12 chunks of code context, which outperformed other models in METEOR score (44.08). This result suggests that repository-level code summarization benefits from few-shot examples, leading to improved performance. However, the effectiveness of few-shot examples is critical, as context chunks only did not ensure shorter summaries. The Modified CodeSearchNet results further underscore the value of few-shot learning in guiding LLMs for code summarization beyond the function level.

Unexpectedly, the largest \textit{deepseek-coder-33b} model performed poorly, suggesting a tendency to produce overly detailed summaries, contrary to our goal of concision. Conversely, non-code-specific \textit{llama3-8b} model, performed well in the SIDE metric, indicating strong semantic alignment between the generated summaries and the corresponding code. This performance warrants a reevaluation of the SIDE metric and deeper analysis of the generated summaries, which we plan to explore in future work.

In summary, the combined results from both datasets indicate that few-shot examples significantly enhance the performance of LLMs in code summarization. While baseline models perform well due to specialized fine-tuning, LLMs demonstrate remarkable adaptability when provided with appropriate guidance. The findings highlight the potential of LLMs to effectively summarize code at various levels, provided that such models are equipped with high-quality examples and relevant context.

\vspace*{-1mm}
\section{Related Work}
\vspace*{-1mm}

The literature on code summarization has predominantly focused on function-level summaries, utilizing encoder-decoder architectures and Transformer-based models to generate concise descriptions of individual functions \cite{vaswani2017attention}, \cite{CodeT5}. These methods have significantly successed in capturing the essence of code snippets but often overlooked the broader context provided by classes and repositories, which is essential for understanding complex codebases. Recent studies have begun to explore code summarization at higher levels, leveraging techniques such as RAG and few-shot learning to incorporate additional context \cite{StarCoder}, \cite{liu2023repobench}. In addition, \cite{su2024revisiting} explored important research questions regarding file context in code summarization. However, the field still lacks comprehensive evaluation of code summarization models at the class and repository levels. What is more, existing benchmarks fail to assess performance beyond the function level.

LLMs have shown promise in various NLP tasks due to their ability to capture long-range dependencies and contextual information \cite{GPT-4}, \cite{DeepSeekCoder}. In code summarization, LLMs with in-context learning capabilities have the potential to generate more informative summaries by leveraging additional context from class and repository structures. Previous work has explored project-specific code summarization using few-shot examples and neural prompt selection \cite{PSCodeSum}, but the application of LLMs in this setting remains underexplored. This study builds upon the above advancements by evaluating the effectiveness of LLMs in code summarization beyond the function level, addressing the existing research gap and contributing to the development of models that generate more contextually relevant, concise, and accurate summaries.

\vspace*{-1mm}
\section{Limitations and Future Work}
\vspace*{-1mm}

Despite the promising findings, this study has several limitations. First, no established benchmarks exist for code summarization beyond the function level, which hinders direct comparison with the existing approaches and complicates the evaluation. Second, the existing datasets are of subpar quality, which may have affected the training and assessment of the models, potentially limiting their generalizability. Third, Naive RAG pipeline presents limitations in effectively leveraging repository context, which may have impeded performance improvements at the repository level. Naive RAG did not help as much as expected; therefore, we leave the manual investigation of this question for future work. Lastly, prompt engineering is time-consuming, especially when crafting effective few-shot examples. This limitation poses scalability challenges for practical applications. To address the above limitations, we plan to explore promising advanced RAG methods in future work to enchance context utilization and improve code summarization. The examples of such edvanced methods are RAG utilizing knowledge graphs such as GraphRAG \cite{GraphRAG} or autonomous LLM agents such as AriGraph \cite{AriGraph}.

\vspace*{-1mm}
\section{Conclusion}
\vspace*{-1mm}

This study demonstrated that code summarization models could effectively operate beyond the function level, particularly when enhanced with additional context and few-shot learning. Fine-tuned language models such as CodeT5+ exceled in code summarization tasks, while incorporating few-shot examples significantly boosted the performance of LLMs such as DeepSeek Coder and Starcoder2. Although repository-level summarization exhibited potential—evidenced by models like DeepSeek Coder—such summarization demanded substantial computational resources and benefited more from structured context than mere repository data. This study lays a foundation for further advancements in the field by providing insights into optimizing code summarization at various levels and highlighting the need for more comprehensive, diverse, and representative benchmarks and datasets. While our study has certain limitations, it represents the first comprehensive exploration of using class- and repository-level context in code summarization, paving the way for future research.

\vspace*{-1mm}
\section*{Acknowledgment}
\vspace*{-1mm}

All authors were supported by the Research Center of the Artificial Intelligence Institute of Innopolis University.


\bibliographystyle{./IEEEtran}
\vspace*{-1mm}
\bibliography{./bib}

\end{document}